\documentclass{ceurart}

\usepackage[frozencache=true,cachedir=minted-cache]{minted}
\setminted{breaklines=true}

\graphicspath{{./}}

\usepackage{subcaption}

\begin{document}

\copyrightyear{2023}
\copyrightclause{Copyright for this paper by its authors.
	Use permitted under Creative Commons License Attribution 4.0
	International (CC BY 4.0).}

\conference{IberLEF 2023, September 2023, Jaén, Spain}

\title{UPB at IberLEF-2023 AuTexTification: Detection of Machine-Generated Text using Transformer Ensembles}

\author[1]{Andrei-Alexandru Preda}[%
	email=andrei.preda3006@stud.acs.upb.ro,
]

\author[1]{Dumitru-Clementin Cercel}[%
	email=dumitru.cercel@upb.ro,
]

\author[1]{Traian Rebedea}[%
	email=traian.rebedea@upb.ro
]

\author[1]{Costin-Gabriel Chiru}[%
	email=costin.chiru@upb.ro,
]

\address[1]{Computer Science and Engineering Department, Faculty of Automatic
	Control and Computers, University Politehnica of Bucharest, Bucharest 060042, Romania}

\begin{abstract}
This paper describes the solutions submitted by the UPB team to the AuTexTification shared task, featured as part of IberLEF-2023. Our team participated in the first subtask, identifying text documents produced by large language models instead of humans. The organizers provided a bilingual dataset for this subtask, comprising English and Spanish texts covering multiple domains, such as legal texts, social media posts, and how-to articles. We experimented mostly with deep learning models based on Transformers, as well as training techniques such as multi-task learning and virtual adversarial training to obtain better results. We submitted three runs, two of which consisted of ensemble models. Our best-performing model achieved macro F1-scores of $66.63\%$ on the English dataset and $67.10\%$ on the Spanish dataset.
\end{abstract}


\maketitle

\section{Introduction}

Recently, computer-generated content started growing in presence on the Internet. With the public release of powerful Large Language Models (LLMs) such as the Generative Pre-trained Transformer~\citep{radford2018improving}, and its derivative systems such as ChatGPT~\citep{team2022chatgpt}, manufacturing texts is easier than ever and probably harder to detect than ever. This phenomenon has already raised several ethical issues that society must answer soon. This effort can be helped by finding mechanisms to automatically and reliably detect computer-generated text.

The AuTexTification: Automated Text Identification shared task~\citep{autextification} is a natural language processing (NLP) competition at IberLEF-2023~\citep{overviewIberLEF2023}. Its main focus is detecting and understanding the computer-generated text, especially that of LLMs. The competition presents two subtasks: (1) Subtask 1 is a binary classification problem in which we have to detect whether a human or an artificial intelligence model wrote a document, and (2) Subtask 2 is a multi-class classification problem in which you have to select which LLM generated a given document from a list of several LLMs. To address these subtasks, the organizers made available a bilingual dataset of documents produced by humans and computers in English and Spanish, covering several domains.

Our team participated only in the first subtask. %
We first experimented with more standard machine learning methods before moving to deep learning models, where we explored techniques such as multi-task learning (MTL)~\cite{collobert2008unified} and virtual adversarial training (VAT)~\cite{miyato2015distributional}. Finally, we combined multiple models trained independently to form ensembles, which we used to generate our submissions, since they performed the best.

\section{Related Work}

While text classification is one of NLP's fundamental and most well-established tasks, detecting computer-generated text is a relatively novel task. This is probably because, until recently, few systems could produce text realistic enough to fool humans. Creating such texts is commonly called natural language generation~\citep{paris2013natural}.

Currently, there seem to be various ways of addressing this problem, which can be classified into black-box and white-box methods~\citep{tang2023science}. White-box techniques require access to the target language model, and they can involve concepts such as watermarks which the models could embed into their outputs to make detection easier. As such, black-box methods are more relevant to the previously mentioned task since we only have access to the model's output, but we do not even know which model produced it.

Black-box methods can involve both classical machine learning classification algorithms, as well as ones based on deep learning~\citep{tang2023science}. To make predictions, traditional algorithms combine statistical features and linguistic patterns with classifiers such as Support Vector Machines (SVMs)~\cite{hearst1998support}. On the other hand, deep learning methods usually involve fine-tuning pre-trained language models using supervised learning in order to make predictions. These deep learning approaches often obtain state-of-the-art results but are harder to interpret, which means they are also harder to trust, as well.

\section{Methods}

This section describes the different classification methods we tried and the final ensemble architectures we submitted.

\subsection{Shallow Learning Models}

\subsubsection{Readability Scores}

Similar to~\citet{stodden2021rs_gv}, we combined several linguistic features with pre-trained embeddings. Specifically, we computed the following readability scores: the Flesch reading ease score~\citep{flesch1948new}, the Gunning-Fog index~\citep{gunning1952technique}, and the SMOG index~\citep{mc1969smog}. The intuition behind this choice was that LLMs might not consider the ease of comprehension when generating texts. For example, the generated legal texts might be harder to understand than those written by humans. To compute the aforementioned  scores, we used the Readability Python library~\citep{readability-library}, which offered $35$ such features.

Then, we concatenated the readability scores with document-level pre-trained embeddings offered by the spaCy
library~\citep{Honnibal_spaCy_Industrial-strength_Natural_2020}, which are $300$-dimensional and language-specific. The English embeddings are based on GloVe~\citep{pennington2014glove}, while the Spanish ones are based on FastText~\citep{mikolov2018advances}. Finally, all features were scaled to have zero mean and unit variance with scikit-learn's \mintinline{python}{StandardScaler}~\citep{scikit-learn}, before using them to train two classifiers, namely XGBoost~\citep{Chen:2016:XST:2939672.2939785} and k-Nearest Neighbors (kNN)~\citep{scikit-learn}.

\subsubsection{String Kernels}

We also experimented with string kernels~\citep{gaman-2023-using}, which are kernel functions that measure the degree of similarity between two strings. An example of a simple string kernel counts the number of n-grams shared by the two strings without considering duplicates. Such a function can be computed for multiple sizes of n-grams, and used as the kernel function of a classifier such as an SVM.

We performed common natural language preprocessing operations on the input text: removing punctuation, removing stopwords, lowercasing all letters, and stemming the words. We used n-gram sizes between $3$ and $5$, and the SVM classifier implemented in scikit-learn~\cite{scikit-learn}. Since custom kernels might need to be computed between each pair of input samples, using the entire training dataset would have taken a long time, so we tested the method only on a small slice of it, comprising several thousand samples.

\subsection{Deep Learning Models}

\subsubsection{Transformers}\label{sec:transformers-mlp}

The Transformer architecture was introduced in 2017 by~\citet{vaswani2017attention} and is currently powering numerous state-of-the-art solutions for many tasks. Transformers usually feature two main components: an encoder and a decoder. However, these two parts can be useful by themselves as well. One example is the Bidirectional Encoder Representations from \mbox{Transformers}~(BERT)~\citep{devlin2018bert} model family, which can encode input text into contextual embeddings.

We experimented with several BERT versions: multilingual ones (i.e., XLM-RoBERTa~\citep{conneau2019unsupervised} and multilingual BERT~\cite{devlin2018bert}), and one pre-trained on tweets (i.e., TwHIN-BERT~\cite{zhang2022twhin}). Since BERT models can be large and typically require large amounts of data to train from scratch, we utilize transfer learning~\citep{weiss2016survey} instead, by fine-tuning pre-trained models.

We experimented with Transformer-based models to encode the raw input text into embeddings, which we then connected to a linear layer, followed by a dropout layer~\cite{srivastava2014dropout} before the final prediction head. The last layer produces a probability of the document being computer-generated, and the binary prediction is chosen by comparing it with a threshold.

\subsubsection{Multi-Task Learning}

As an additional method of preventing overfitting, we used the technique of multi-task learning. MTL refers to training a model to solve multiple tasks simultaneously. As such, these models typically feature a set of parameters shared for all tasks, and separate prediction heads for each task. Intuitively, multi-task learning should make the task harder to solve, thus adding extra complexity, which the model has to adapt to.

In our case, an extra task that is easy to derive is predicting the language of a given input document. More precisely, apart from predicting the human/computer label of a document, the model has to detect whether it is written in English or Spanish. This means that, for training, we combined the two datasets supplied for Subtask 1. However, we did not use any of the data provided for Subtask 2.

The MTL architecture is very similar to the one presented in the previous section, only adding an extra classification head. The architecture can be seen in Figure~\ref{fig:mtl}. Since both tasks involve binary classification, we compute a binary cross-entropy loss for each of them, namely $\mathcal{L}_{\mathrm{bot}}$ for the human/computer classification task, and $\mathcal{L}_{\mathrm{lang}}$ for the language detection task. The final loss of the model is a combination of these two losses, as given by the following formula:
\begin{equation}
    \mathcal{L} = \alpha \mathcal{L}_{\mathrm{bot}} + (1-\alpha) \mathcal{L}_{\mathrm{lang}}
\end{equation}
where hyperparameter $\alpha$ controls how much attention is paid to each task.

\subsubsection{Virtual Adversarial Training}

VAT~\citep{miyato2018virtual} is another regularization technique for deep learning models. It aims to help models generalize better by perturbing the inputs to maximize the loss function. For our models, the inputs refer to the embeddings of the raw documents, not to the token IDs.

In our case, this method implies performing the forward and backward passes multiple times in order to compute the gradients. Then, the loss function specific to VAT gets added to the regular loss function, the final loss being the sum of the two. We added VAT to our models using the VAT-pytorch Python library~\citep{VAT-pytorch}, which implements the distributional smoothing technique described by~\citet{miyato2015distributional}.

\subsection{Ensemble Learning}
Ensemble techniques combine multiple different models to make better predictions. Intuitively, they should make the weaknesses of each model matter less since if one model happens to have poor performance on a certain edge case, all the other models will probably give better results, thus negating the impact of the incorrect prediction.

\begin{figure}
	\centering
	\begin{subfigure}{.29\textwidth}
		\centering
		\includegraphics[width=\linewidth]{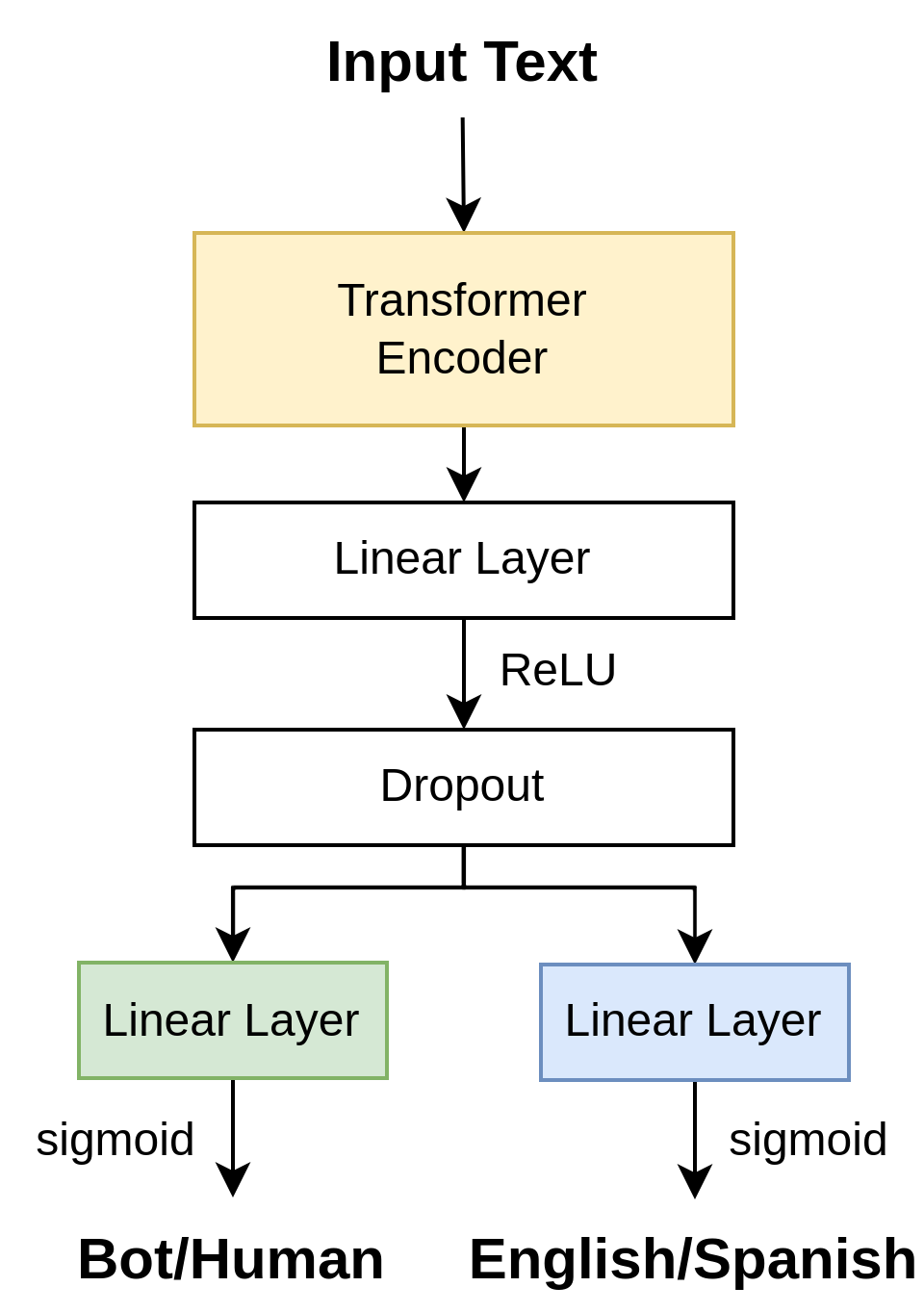}
		\caption{The MTL \mbox{architecture} used for
			\mintinline{text}{run 1}.}
		\label{fig:mtl}
	\end{subfigure}
	\begin{subfigure}{.69\textwidth}
		\centering
		\includegraphics[width=\linewidth]{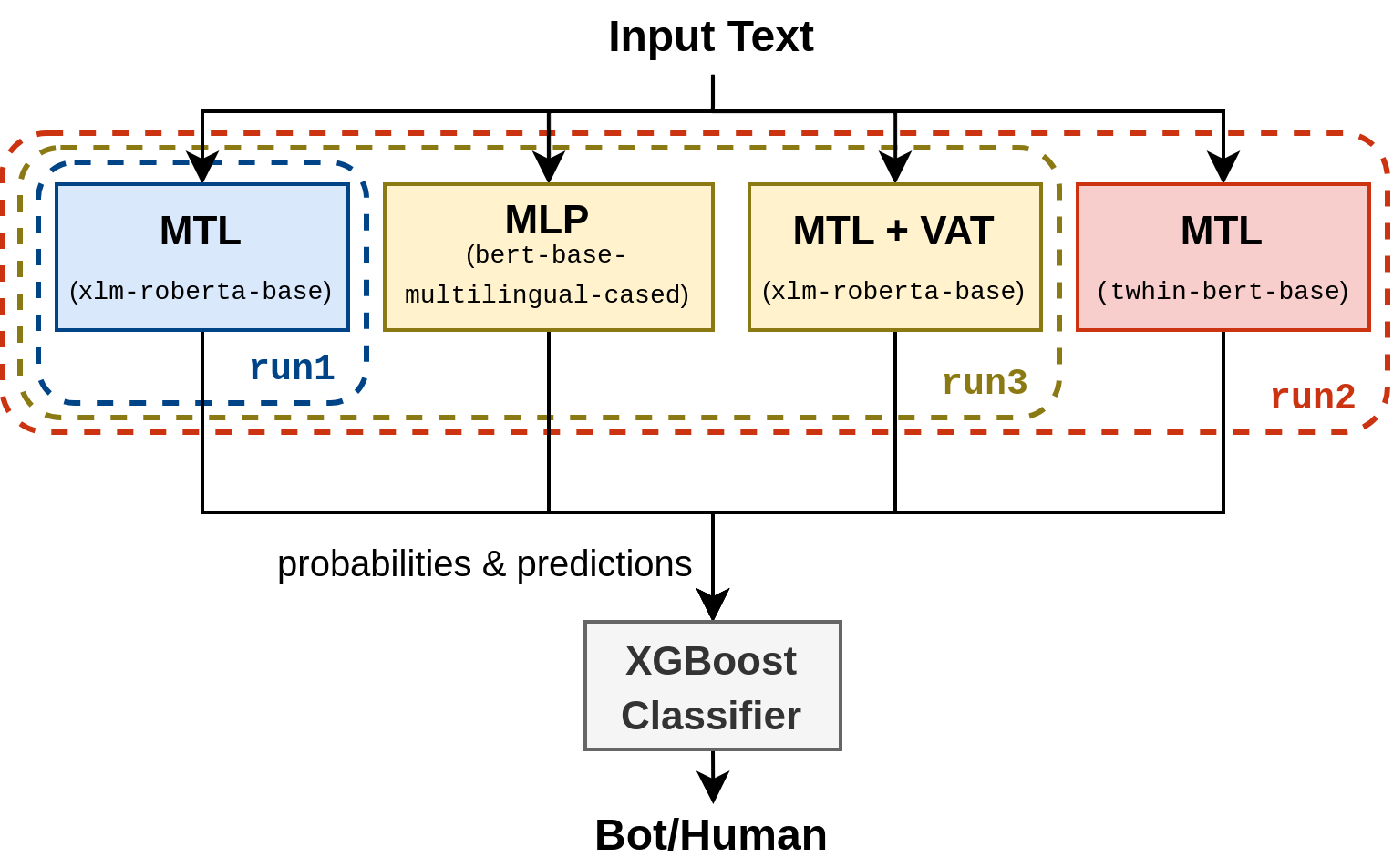}
		\caption{Summary of the ensemble architectures used for runs 2 and 3.}
		\label{fig:ensembles}
	\end{subfigure}
	\caption{The architectures used to create the submissions.
	\mintinline{text}{run1} featured an MTL model, while the other two runs used ensembles. The difference between
    \mintinline{text}{run2} and \mintinline{text}{run3} is that the latter misses the model pre-trained on Tweets. The Multilayer Perceptron (MLP) architecture is described in Subsection~\ref{sec:transformers-mlp}.}
	\label{fig:architectures}
\end{figure}

While there are multiple ways of combining models into ensembles (such as majority voting or bagging)~\cite{galar2011review}, we decided to use the stacking technique inspired by the work of~\citet{gaman-2023-using}. Thus, we train an extra meta-learner model, which learns to make predictions based on the outputs of each model in the ensemble. We experimented with an XGBoost classifier, which takes as input the probabilities produced by each model, and the binary predictions they make.
The submitted final ensembles can be seen in Figure~\ref{fig:ensembles}.

\section{Experiments}

\subsection{Dataset}

The training dataset provided for Subtask 1 consists of approximately $33{,}845$ English documents, and $32{,}062$ Spanish documents. For both languages, the ratio of computer-generated to human-generated texts was roughly $50\%$. This suggested that the dataset we worked with was fairly well-balanced, and we did not attempt to use any techniques for dealing with imbalanced data.

While the individual BERT classifier (MLP) was trained for the two languages separately, the multi-task learning experiments merged the two slices into a single dataset. In both cases, we used $70\%$ of the labeled data to train the models, and we set aside the other $30\%$ for validation. This split was fixed at the beginning of the project to allow us to compare the models' performance. However, when creating the final predictions, we retrained all models on the full labeled dataset, only using $2\%$ of it (around $1{,}300$ samples) as a validation set to monitor the training process.

\begin{figure}
	\centering
	\includegraphics[width=0.95\linewidth]{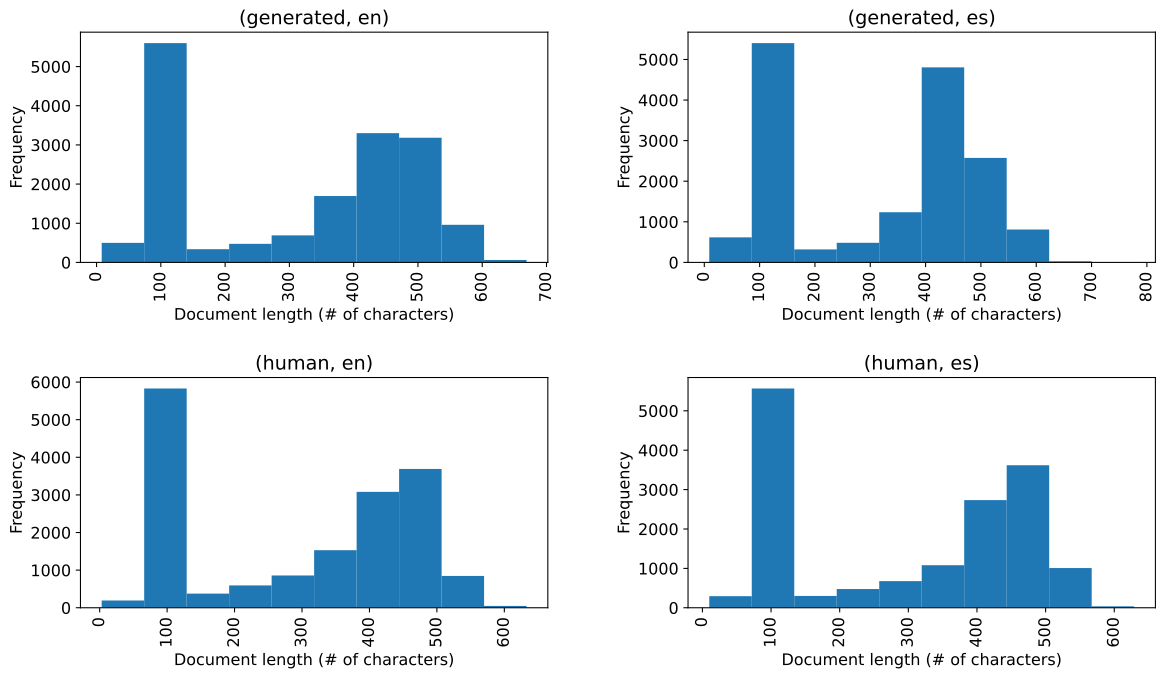}
	\caption{The length distribution of the documents in the training set, grouped by label and language.}
	\label{fig:length-dist}
\end{figure}

As seen in Figure~\ref{fig:length-dist}, there is a relatively large number of documents with fewer than $200$ characters. Since the task organizers described one of the domains in the dataset as social media, and Twitter usually limits user posts to approximately $300$ characters at most, we assumed these samples to be tweets. For this reason, we experimented with a language model pre-trained specifically on tweets, namely TwHIN-BERT.

We did not perform any data preprocessing for the deep learning models, using the raw documents as input instead.

\subsection{Hyperparameters}

For the deep learning models, we used a hidden layer of size $64$, and a dropout rate of $0.2$. For MTL, we assigned the same weight to the two loss values, i.e., $\alpha = 0.5$. We used the default values for VAT, namely $\alpha = 1$, $\epsilon = 1$, and $\xi = 10$.

One of the parameters we did not set to a fixed value was the threshold for turning the probabilities into binary predictions. To choose this threshold, we compute the true positive rate (TPR) and false positive rate (FPR) of the Receiver Operating Characteristic curve for the validation set. Since the dataset was balanced, we did not want to favor either precision or recall, so we picked the threshold which brings the sum of TPR and FPR closest to $1$. In our experiments, the threshold was often greater than $0.9$, sometimes over $0.95$.

We used the AdamW optimizer~\cite{kingma2017adam} implemented in PyTorch~\cite{pytorch} with a learning rate of $10^{-5}$, and $2$-$4$ epochs for training. In order to avoid overfitting, we used both a dropout layer and the early stopping technique. We used batch sizes between $24$ and $48$, depending on the model size.

For the kNN shallow model, we set the parameter $k$ to $10$. For the final ensembles, we performed a grid search to find the hyperparameters of the XGBoost model, which maximized the macro F1-score on a small validation set for each of the two languages. We searched for estimators between $2$ and $30$, depths between $3$ and $10$, and learning rates between $10^{-5}$ and $10^{-1}$. The best hyperparameters were: $\mathrm{n\_estimators} = 3$, $\mathrm{max\_depth} = 5$, and $\mathrm{learning\_rate} = 10^{-3}$.

\subsection{Results}

The results obtained for Subtask 1 can be seen in Tables~\ref{tab:results-en} and~\ref{tab:results-es}, alongside the baselines provided by the task organizers, the best results obtained by other participant teams, and our other experiments, which we did not submit.

\begin{table*}
	\caption{Macro-F1 (F1) scores obtained on the English dataset for Subtask 1. Our experiments are highlighted in bold. The validation set refers to the one we used, while the organizers provided the test set.}
	\label{tab:results-en}
	\begin{tabular}{lccc}
		\toprule
		Model & Rank & Validation Set F1 & Test Set F1 \\
		\midrule
		TALN-UPF Hybrid Plus
		      & 1    & -             & 80.91 \\
            TALN-UPF Hybrid
		      & 2    & -             & 74.16 \\
		\textbf{Full ensemble (our \mintinline{text}{run2})}
		      & 18   & -             & 66.63 \\
		\textbf{Ensemble without TwHIN-BERT (our \mintinline{text}{run3})}
		      & 19   & -             & 66.40 \\
		Logistic Regression (baseline)
		      & 23   & -             & 65.78 \\
		\textbf{MTL (\mintinline{text}{xlm-roberta-base}) (our \mintinline{text}{run1})}
		      & 25   & 93.30         & 65.53 \\
		Symanto Brain (Few-shot) (baseline)
		      & 37   & -             & 59.44 \\
		DeBERTa V3 (baseline)
		      & 51   & -             & 57.10 \\
		Random (baseline)
		      & 69   & -             & 50.00 \\
		Symanto Brain (Zero-shot) (baseline)
		      & 73   & -             & 43.47 \\

		\hline

		\textbf{MTL (\mintinline{text}{bert-base-multilingual-cased})}
		      & -    & 92.70         & - \\
		\textbf{MLP (\mintinline{text}{bert-base-multilingual-cased})}
		      & -    & 91.80         & - \\
		\textbf{XGBoost + Readability + GloVe}
		      & -    & 79.80         & 59.22 \\
		\textbf{kNN + Readability + GloVe}
		      & -    & 74.90         & 56.31 \\
		\bottomrule
	\end{tabular}
\end{table*}

\begin{table*}
	\caption{Macro-F1 (F1) scores obtained on the Spanish dataset for Subtask 1. Our experiments are highlighted in bold. The validation set refers to the one we used, while the organizers provided the test set.}
	\label{tab:results-es}
	\begin{tabular}{lccc}
		\toprule
		Model & Rank & Validation Set F1 & Test Set F1 \\
		\midrule
		TALN-UPF Hybrid Plus
		      & 1    & -             & 70.77 \\
            Linguistica\_F-P\_et\_al
		      & 2    & -             & 70.60 \\
		RoBERTa (BNE) (baseline)
		      & 3    & -             & 68.52 \\
		\textbf{Ensemble without TwHIN-BERT (our \mintinline{text}{run3})}
		      & 6    & -             & 67.10 \\
		\textbf{Full ensemble (our \mintinline{text}{run2})}
		      & 7    & -             & 66.97 \\
		\textbf{MTL (\mintinline{text}{xlm-roberta-base}) (our \mintinline{text}{run1})}
		      & 12   & 92.30         & 65.01 \\
		Logistic Regression (baseline)
		      & 25   & -             & 62.40 \\
		Symanto Brain (Few-shot) (baseline)
		      & 39   & -             & 56.05 \\
		Random (baseline)
		      & 46   & -             & 50.00 \\
		Symanto Brain (Zero-shot) (baseline)
		      & 50   & -             & 34.58 \\

		\hline

		\textbf{MTL (\mintinline{text}{bert-base-multilingual-cased})}
		      & -    & 91.00         & - \\
		\textbf{MLP (\mintinline{text}{bert-base-multilingual-cased})}
		      & -    & 90.90         & - \\
		\textbf{XGBoost + Readability + FastText}
		      & -    & 80.90         & 63.70 \\
		\textbf{kNN + Readability + FastText}
		      & -    & 72.20         & 59.59 \\
		\bottomrule
	\end{tabular}
\end{table*}

As expected, the ensembles performed better than the MTL model (i.e., \mintinline{text}{run1}) on both datasets. However, the scores obtained on the test set are much smaller than those obtained on our validation set. This fact could indicate that our chosen validation set was too small or poorly chosen, or that the distribution of the test set is different from that of the training set. It would have been interesting to see if methods such as cross-validation would have produced other validation scores closer to the real performance.

Our experiments indicated that the choice of Transformer mattered as well, with multilingual models being generally better for this use case. Similarly, the fine-tuned embeddings produced during training were better than the pre-trained ones provided by spaCy, and hence, we did not explore the shallow learning direction more.

We did not perform a grid search or any other type of hyperparameter search for the deep learning models, so our approaches could obtain better results simply by choosing appropriate hyperparameter values. Another thing to note is that while the full ensemble performed better on the English dataset, removing the TwHIN-BERT Transformer slightly improved the results on the Spanish dataset.

\section{Conclusions}

In this paper, we proposed multiple methods for addressing the task of detecting LLM-generated text. While we experimented briefly with more classical machine learning classifiers, we saw that Transformer-based models performed better for this task. We described our experiments with several classification models powered by BERT and detailed some regularization techniques, which improved their performance slightly. Finally, we stacked multiple such models to form ensembles, leading to even better performance and achieving our best macro F1-scores of 66.63\% on the English dataset, and $67.10\%$ on the Spanish dataset of the AuTexTification shared task, Subtask 1.

Regarding future work,  we could improve the choice of hyperparameters since they are often crucial for achieving good performance, and techniques such as grid search should find better values. Similarly, the training process can be improved by using better methods to avoid overfitting, and increasing the number of training epochs.

\bibliography{bibliography}

\appendix

\end{document}